\documentclass[conference]{IEEEtran}
\IEEEoverridecommandlockouts
\usepackage{cite}
\usepackage{amsmath,amssymb,amsfonts}
\usepackage{algorithmic}
\usepackage{graphicx}
\usepackage{textcomp}
\usepackage{xcolor}
\usepackage{bm}
\usepackage{tikz}
\usepackage{float}
\usepackage{comment}
\usepackage{booktabs}
\usepackage{url} 
\usepackage[caption=false,font=footnotesize]{subfig}
\usetikzlibrary{arrows.meta,positioning,calc}

\newcommand{\cellimg}[2]{\subfloat[#2]{\includegraphics[height=2.5cm,width=1.6cm]{#1}}}

\def\BibTeX{{\rm B\kern-.05em{\sc i\kern-.025em b}\kern-.08em
    T\kern-.1667em\lower.7ex\hbox{E}\kern-.125emX}}
\begin{document}

\title{Decentralized Multi-Robot Obstacle Detection and Tracking in a Maritime Scenario\\
}

\author{\IEEEauthorblockN{Muhammad Farhan Ahmed}
\IEEEauthorblockA{\textit{ARMEN Team} \\
\textit{LS2N, Ecole Centrale de Nantes (ECN)}\\
Nantes, France \\
Muhammad.Ahmed@ec-nantes.fr}
\and
\IEEEauthorblockN{Vincent Frémont}
\IEEEauthorblockA{\textit{ARMEN Team} \\
\textit{LS2N, Ecole Centrale de Nantes (ECN)}\\
Nantes, France \\
vincent.fremont@ec-nantes.fr}
}

\maketitle

\begin{abstract}
Autonomous aerial–surface robot teams offer a scalable solution for maritime monitoring, but deployment remains difficult due to water-induced visual artifacts and bandwidth-limited coordination. This paper presents a decentralized multi-robot framework to detect and track floating containers using multiple UAVs cooperating with an autonomous surface vessel. Each UAV runs a YOLOv8 detector augmented with stereo disparity and maintains per-target EKF tracks with uncertainty-aware data association. Robots exchange compact track summaries that are fused conservatively using Covariance Intersection, preserving estimator consistency under unknown cross-correlations. An information-driven allocator assigns targets and selects UAV hover viewpoints by trading expected uncertainty reduction in travel effort and safety separation. Implemented in ROS, the proposed system is validated in simulations and compared with representative tracking and fusion baselines, showing improved identity continuity and localization accuracy with modest communication overhead.
\end{abstract}

\begin{IEEEkeywords}
Multi-robot systems, Maritime perception, Covariance intersection, Multi-robot task allocation
\end{IEEEkeywords}

\section{Introduction}
\label{sc:introduction}
Teams of aerial and surface robots are increasingly used for maritime inspection, monitoring, and search-and-rescue, but over-water operation is challenging: reflections, waves, and low texture degrade vision, while long ranges and intermittent line-of-sight limit communication. Reliable deployment therefore requires accurate state estimation, robust perception, and scalable coordination under limited bandwidth~\cite{paull2014auv}.

In this article, we present a decentralized multi-robot perception framework for detecting and tracking floating containers using multiple UAVs cooperating with an autonomous surface vessel. Each agent runs onboard localization, and UAVs use stereo perception to obtain metric 3D container observations. Targets are tracked locally, while robots exchange compact track summaries that are fused conservatively to remain consistent under unknown cross-correlations. Based on the fused tracks, an information-driven allocator assigns targets and selects UAV hover viewpoints by trading expected uncertainty reduction against travel effort and inter-UAV separation.

Our main contributions are:
(i) a decentralized maritime multi-robot detection and tracking pipeline that combines stereo-based 3D perception, EKF tracking, and conservative fusion via Covariance Intersection;
(ii) an information-driven target allocation method that prioritizes targets by expected uncertainty reduction while enforcing safety and workload constraints; and
(iii) a mode-switching and hover-viewpoint selection strategy that improves observability, coverage, and tracking accuracy.

This paper is organized as follows. Section~\ref{sc: related_work} reviews related work. Section~\ref{sec:methodology} presents the proposed method. Section~\ref{sc sim-results} describes the simulation setup and results. Section~\ref{sc: conclusions} concludes and outlines directions for future work.

\section{RELATED WORK}
\label{sc: related_work}
\subsection{Maritime Visual Detection}\label{sc: Maritime_Visual_Detection}
Recent real-time detectors increasingly favor end-to-end training to improve the latency--accuracy trade-off~\cite{minh1}. 
However, maritime UAV imagery exhibits substantial domain shift (e.g., reflections, haze, low horizon contrast, and large-scale variation), motivating maritime-specific training and adaptation for robustness~\cite{GENG2, mm1}. 
Public benchmarks such as SeaDronesSee and related maritime vision challenges enable systematic evaluation in open-water conditions~\cite{varga2022seadronessee, macvi2023seadronesseecvprw, real1, mota1}. 
In our pipeline, we fuse detector inputs with stereo disparity to obtain metric 3D observations, which improves data association under occlusion. 
We build on recent learning-based matching methods that remain robust in low-texture and long-range regimes~\cite{li2022practicalstereo}.

\subsection{Multi-Object Tracking}\label{sc: Multi-Object_tracking}
Tracking pipelines increasingly emphasize stronger association and more stable motion models, with recent baselines improving robustness in crowded scenes and under detection noise~\cite{zhang2022botsort}. 
In 3D, simple Bayesian filtering can remain competitive when measurements are metric and uncertainty is well modeled~\cite{weng2020ab3dmot}. 
For maritime tracking, lightweight trackers are particularly attractive due to their stability under intermittent observations and modest compute constraints~\cite{shao2024robust}. 
Recent maritime datasets also expose failure modes induced by glare, wake effects, and rapid appearance changes~\cite{bakht2025mvtd}. 
Motivated by these findings, we adopt independent EKFs with Mahalanobis gating and use conservative inter-robot fusion to avoid overconfident cross-agent track updates.

\subsection{Decentralized Fusion}\label{sc: Decentralized_Fusion}
Multi-robot estimation must account for unknown inter-robot correlations arising from shared priors, repeated communication, and common sensing. 
Recent work on consistent distributed cooperative localization therefore, emphasizes conservative fusion and estimator consistency under realistic networking constraints~\cite{zhou2024dinekf, chang2021resilientci, global1}. 
In this spirit, we apply track-level Covariance Intersection (CI) with gating to avoid fusing unrelated hypotheses and to remain robust under intermittent communication~\cite{julierCI}.

\subsection{Assignment and Task Allocation}\label{sc: Assignment_and_Task_Allocation}
Dynamic multi-robot task allocation (MRTA) has gained renewed attention through optimization-based formulations and scalable solvers. 
Recent surveys summarize core optimization techniques and provide practical guidance on modeling costs, constraints, and dynamics in MRTA~\cite{chakraa2023mrta_review}. 
For dynamic scenarios, recent work explores online and formal optimization approaches that better capture feasibility and evolving task demands~\cite{freitas2024smt_mrta, gab2025aamas_mrta}. 
In our maritime perception setting, we adopt a capacitated min-cost flow formulation, with costs that combine expected information gain, travel effort, safety separation, and assignment stickiness.

\section{METHODOLOGY}
\label{sec:methodology}
Existing maritime pipelines often rely on 2D detections and online association, which are fragile over reflective water and low contrast, causing noisy measurements and fragmented tracks~\cite{detect1, wang2025srdetr, wojke2017deepsort}. In multi-robot settings, centralized fusion also increases bandwidth dependence and can become inconsistent under correlated estimates, leading to overconfident covariances~\cite{wu2018consistencyfusion, tu2025wci}. 

Our method addresses these issues and keeps perception and tracking local, exchanges only compact summaries, fuses tracks conservatively via CI, and couples multi-target assignment with uncertainty-aware viewpoint selection.
We first introduce the notation and system overview, then describe localization, detection, tracking, fusion, assignment, hover selection, and mode management. 
Throughout the paper, we use ‘UAV’ and ‘drone’ interchangeably to denote aerial agents, and ‘container’ and ‘target’ interchangeably to denote tracked floating objects.
\subsection{System Overview and Notation}
\label{sec:system_overview}
We consider multiple UAVs and one autonomous surface vessel (Aquabot) operating in a world frame $w$.
Each UAV $j\in\{1,\dots,M\}$ has position $\bm r_j\in\mathbb{R}^3$.
Each container/target $i\in\{1,\dots,N\}$ is represented by a fused 3D position estimate
$(\hat{\bm p}_i, P_i)$, where $\hat{\bm p}_i\in\mathbb{R}^3$ and $P_i\in\mathbb{R}^{3\times 3}$.

\begin{figure}[t]
\centering
\scriptsize
\begin{tikzpicture}[
  node distance=3mm and 7mm,
  >=stealth,
  block/.style={
    rectangle,
    draw,
    rounded corners=1pt,
    thick,
    align=center,
    minimum height=4.5mm,
    inner sep=2pt,
    fill=green!30
  },
  bigblock/.style={
    rectangle,
    draw,
    rounded corners=1pt,
    thick,
    align=center,
    minimum height=5mm,
    inner sep=2pt,
    fill=blue!30
  },
   modemode/.style={
    rectangle,
    draw,
    rounded corners=1pt,
    thick,
    align=center,
    minimum height=4.5mm,
    inner sep=2pt,
    fill=green!30   
  },
  line/.style={->, thick}
]

\node[block] (ekf) {EKF Localization\\(GPS+IMU Fusion) \ref{sec:loco}
};
\node[block, right=of ekf] (yolo) {YOLOv8 + Stereo\\3D Detections \ref{sec:yolo-stereo}
};
\node[block, right=of yolo] (mot) {Local MOT EKF\\Per-Object Tracks  \ref{sec:mot}
};

\node[bigblock, below=of yolo] (fuse)
  {Decentralized Fusion and Data Association \\of Fused Tracks \ref{sec:cii} and \ref{sec:data_association}
  };

\node[bigblock, below=of fuse] (assign)
  {CMCF Assignments $\mathcal{A}_j$ for each UAV
  \ref{sec:cmcf}
  };

\node[bigblock, below=of assign] (hover)
  {Hover selection per assigned container \\
   maximizing information gain    
   \ref{sec:hover_pose_selection}   
   };

\node[modemode, below left=9mm and -5mm  of hover] (surv)
  {\textsc{Surveillance}};

\node[modemode, below right=9mm and -5mm of hover] (track)
  {\textsc{Tracking} };

\node[modemode, below=15mm of hover] (done)
  {Check \textsc{Tracking} \\  termination criteria 
    
   \ref{sec:tracking_termination_criteria}};

\draw[line] (ekf) -- (yolo);
\draw[line] (yolo) -- (mot);

\draw[line] (mot.south) |- (fuse.east);
\draw[line] (fuse) -- (assign);
\draw[line] (assign) -- (hover);



\coordinate (split) at ($(hover.south)+(0,-5mm)$);

\draw[line] (hover.south) -- (split);

\draw[line]
  (split) -| node[pos=0.25, above]{$\mathcal{A}_j = \emptyset$}
  (surv.north);

\draw[line]
  (split) -| node[pos=0.25, above]{$\mathcal{A}_j \neq \emptyset$}
  (track.north);

\draw[line]
  (track.south) |- (done.east);

\draw[line]
  (done.west) -| node[pos=0.25, below]{handoff}
  (surv.south);

\end{tikzpicture}
\caption{Pipeline with UAV (green) and Aquabot (blue) nodes.}
\label{fig:method_flow_modes}
\end{figure}
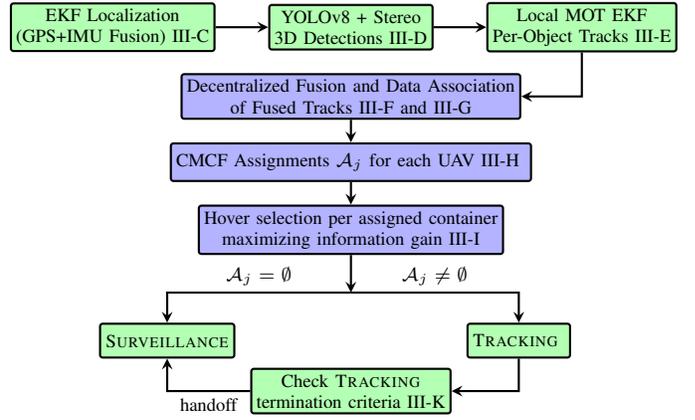

The system is implemented in ROS as shown in Fig.~\ref{fig:method_flow_modes}. Each UAV runs onboard EKF localization leveraging GPS and IMU sensor fusion, YOLOv8+disparity for metric 3D detections, and a per-object EKF tracker with Mahalanobis gating. UAV track summaries are fused on Aquabot using (CI), producing compact fused track arrays published on a shared ROS topic. A dedicated assignment node subscribes to the fused tracks and to all UAV odometry topics. It constructs a cost matrix that encodes a per–UAV–target information-gain, UAV--target distance, stickiness to previous assignments, safety-spacing penalties, and feasibility constraints, and solves a CMCF via successive shortest paths.

Each UAV is issued at least one single \emph{primary} target per allocation cycle. For each assigned UAV--target pair, the node then computes an uncertainty-aware hover pose on a discrete ring and publishes the selected target ID, the hover pose, and the corresponding obstacle state to the downstream navigation components. For each UAV--target pair, the tracking termination criteria are checked, and if satisfied, it enters into $
\textsc{Surveillance}$ mode.  

\begin{figure}[H]
    \centering
    \includegraphics[width=0.70\columnwidth]{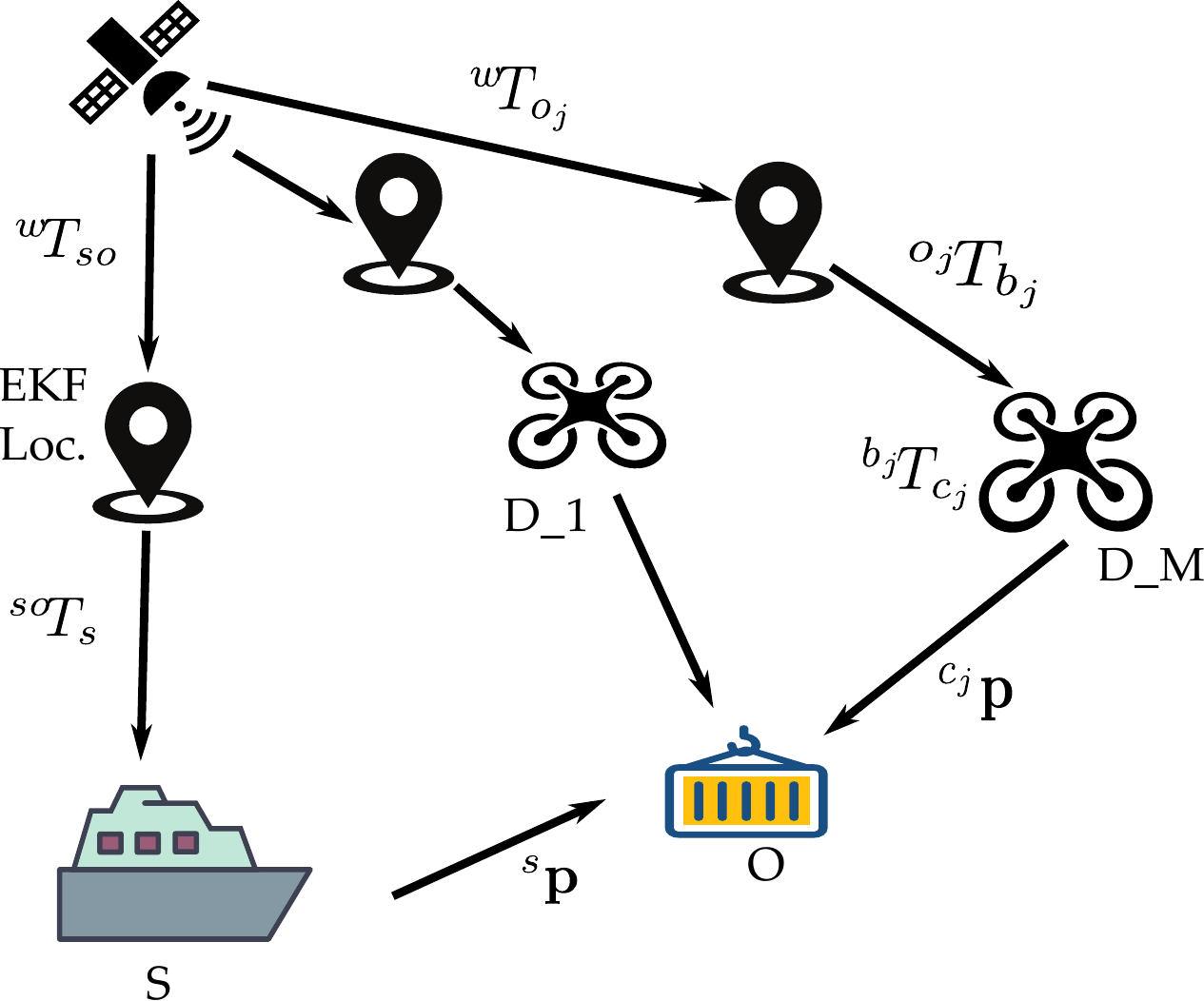}
    \caption{System TFs Overview.}
    \label{fig:system_arch}
\end{figure}
\subsection{Coordinate Frames and Transforms}
\label{sec:frames_tfs}
As shown in Fig.~\ref{fig:system_arch}, all robots share a world frame $w$. Each UAV $j$ uses frames $c_j$ (camera), $b_j$ (body), and $o_j$ (odometry), while Aquabot uses $s$ (base) and $so$ (odometry). We denote the rigid transform from frame $b$ to frame $a$ by ${}^{a}!T_{b}\in\mathbb{R}^{4\times4}$, which maps coordinates from $b$ to $a$.

Given a 3D point expressed in the UAV camera frame ${}^{c_j}\bm p \in \mathbb{R}^3$, we map it to the world frame as
$
{}^{w}\bm p
=
{}^{w}\!T_{o_j}\,{}^{o_j}\!T_{b_j}\,{}^{b_j}\!T_{c_j}
\big[\,({}^{c_j}\bm p)^\top\;\;1\,\big]^\top .
$
Here, ${}^{w}\!T_{o_j}$ is provided by the UAV localization EKF (Sec.~\ref{sec:loco}), while ${}^{o_j}\!T_{b_j}$ and ${}^{b_j}\!T_{c_j}$ are fixed extrinsic transforms in the UAV TF tree.

Similarly, a point in the world frame can be expressed in Aquabot's base frame $s$ using
$
{}^{s}\bm p
=
{}^{s}\!T_{w}\,\big[\,({}^{w}\bm p)^\top\;\;1\,\big]^\top.
$
We compute ${}^{w}\!T_{s}$ as
$
{}^{w}\!T_{s} = {}^{w}\!T_{so}\,{}^{so}\!T_{s},
$
where ${}^{w}\!T_{so}$ is provided by Aquabot's localization EKF (Sec.~\ref{sec:loco}) and ${}^{so}\!T_{s}$ is the fixed transform between Aquabot's odometry and base frames.

Together, these transforms define a consistent pipeline for expressing container measurements in the common world frame and, when needed, in Aquabot's base frame.

\subsection{Agent Localization via EKF}
\label{sec:loco}
Each agent runs an onboard EKF in the shared world frame $w$.
We estimate 3D position, velocity, and acceleration with a constant-acceleration motion model:
$
\bm x_t=\big[\bm p_t^\top,\,\bm v_t^\top,\,\bm a_t^\top\big]^\top \in \mathbb{R}^9 .
\label{eq:ekf_state}
$
The filter predicts at $\Delta t$ and fuses GPS position and IMU gravity-compensated linear acceleration
measurements using standard EKF updates, with conservative noise tuning (increased during aggressive maneuvers).
We initialize from the first valid GPS fix with zero velocity/acceleration and large covariance. The EKF output provides the world-to-odometry transforms ${}^{w}\!T_{o_j}$ (UAVs) and ${}^{w}\!T_{so}$
(Aquabot).

\subsection{YOLOv8 Detection and disparity}
\label{sec:yolo-stereo}
\paragraph{YOLOv8 Detection.}
We adopt early fusion~\cite{wang2022wacv_deepsensorfusion,ophoff_rgbd02} by concatenating rectified disparity with RGB to form a 4-channel input.
We modify YOLOv8 by expanding the first convolution to accept four channels, then fine-tune the network (transfer learning) on five container classes using standard YOLO augmentations. We refer to this model as YOLOv8-RGBDis. The training objective is
\begin{equation}
\mathcal{L}
=
\lambda_{\text{box}}\mathcal{L}_{\text{IoU}}
+
\lambda_{\text{dfl}}\mathcal{L}_{\text{DFL}}
+
\lambda_{\text{cls}}\mathcal{L}_{\text{BCE}},
\label{eq:yolo_loss}
\end{equation}
where $\mathcal{L}_{\text{IoU}}$ is the bounding-box regression loss, $\mathcal{L}_{\text{DFL}}$ is the Distribution Focal Loss, and $\mathcal{L}_{\text{BCE}}$ is the (sigmoid) binary cross-entropy classification loss over the container classes. The non-negative weights $\lambda_{\text{box}},\lambda_{\text{dfl}},\lambda_{\text{cls}}$ balance localization, boundary refinement, and classification.
\paragraph{Stereo 3D via disparity.}
For each detection $k$ with an image-plane bounding box
$\mathcal{B}_k=(u_k, v_k, w_k, h_k)$ and confidence $s_k$, where $(u_k,v_k)$ denotes the bounding-box center in pixels.
We estimate a robust disparity $\hat d_k$ as the median over valid disparity pixels inside the box:
\begin{align}
\hat d_k
&=
\operatorname{median}\!\left\{
d(i,j)\; \middle|\;
(i,j)\in\mathcal{B}_k,\right.\nonumber\\
&\hspace{16mm}\left.
d_{\min}\le d(i,j)\le d_{\max}
\right\}.
\label{eq:median_disp}
\end{align}
where $d(i,j)$ is the rectified disparity value at pixel $(i,j)$ and $[d_{\min},d_{\max}]$ filters invalid/outlier disparities.

Metric depth and camera-frame coordinates follow the pinhole stereo model:
$Z=\frac{fB}{\hat d_k}$, $
X=\frac{(u_k-c_x)Z}{f}$, $
Y=\frac{(v_k-c_y)Z}{f},$ with focal length $f$, principal point $(c_x,c_y)$, and stereo baseline $B$.
We clamp $Z\in[Z_{\min},Z_{\max}]$ and discard detections with insufficient valid disparity support inside $\mathcal{B}_k$.
\subsection{Multi-Object Tracking with EKF}
\label{sec:mot}
Each UAV $j$ maintains a local constant-velocity EKF track for each target $i$,
with state $\hat{\bm x}_{j,i,t}\in\mathbb{R}^6$ and covariance $P^{x}_{j,i,t}\in\mathbb{R}^{6\times 6}$.

Each target is modeled in the world frame with a constant-velocity state:
$
\bm x_{j,i,t} =
\big[\bm p_{j,i,t}^\top,\,\bm v_{j,i,t}^\top\big]^\top \in \mathbb{R}^{6}.
\label{eq:track_state}
$
Stereo provides 3D position observations $\bm z_{j,k,t}\in\mathbb{R}^3$, modeled as
\begin{equation}
\bm z_{j,k,t} = H\,\bm x_{j,i,t} + \bm v_{j,k,t},\quad
\bm v_{j,k,t}\sim\mathcal{N}(0,R_{j,k,t}),
\label{eq:track_meas}
\end{equation}
with
$
H=\begin{bmatrix} I_3 & 0_{3\times 3} \end{bmatrix}.
$
Here, $R_{j,k,t}$ captures range/disparity-dependent uncertainty. Data association uses Mahalanobis gating;
unmatched detections initialize new tracks, and tracks are removed after a timeout or if uncertainty exceeds a threshold. For communication and fusion,
we transmit only the position marginal $(\hat{\bm p}_{j,i,t},P_{j,i,t})$.

\subsection{Decentralized Fusion via Covariance Intersection}
\label{sec:cii}
Track estimates exchanged among robots can exhibit \emph{unknown cross-correlations} due to shared priors, overlapping observations, common process models, and information relaying. Na"ively assuming independence can yield overconfident covariances and inconsistent estimates. To guarantee conservative fusion without cross-covariance terms, we use Covariance Intersection (CI)~\cite{julierCI}.
In our system, each UAV maintains a local 6D constant-velocity EKF per target (Sec.~\ref{sec:mot}), but for communication and fusion we share only the \emph{3D position marginal}
$(\hat{\bm p}_{i,t}, P_{i,t})$.

Consider two robots providing estimates of the same target $i$ at time $t$,
$(\hat{\bm p}^{(1)}_{i,t}, P^{(1)}_{i,t})$ and $(\hat{\bm p}^{(2)}_{i,t}, P^{(2)}_{i,t})$.
CI computes a fused estimate as
\begin{align}
\big(P_{i,t}^{\mathrm{CI}}\big)^{-1} &=
\omega \big(P^{(1)}_{i,t}\big)^{-1} + (1-\omega)\big(P^{(2)}_{i,t}\big)^{-1},
\quad \omega\in[0,1],
\label{eq:ci_cov_short}\\
\hat{\bm p}_{i,t}^{\mathrm{CI}} &=
P_{i,t}^{\mathrm{CI}}
\left(
\omega \big(P^{(1)}_{i,t}\big)^{-1}\hat{\bm p}^{(1)}_{i,t}
+
(1-\omega)\big(P^{(2)}_{i,t}\big)^{-1}\hat{\bm p}^{(2)}_{i,t}
\right).
\label{eq:ci_mean_short}
\end{align}

We choose the mixing weight $\omega$ by minimizing a scalar uncertainty criterion:
$
\omega^\star
=
\arg\min_{\omega\in[0,1]}
\log\det\!\big(P_{i,t}^{\mathrm{CI}}(\omega)\big).
\label{eq:ci_w_short}
$
When more than two robot estimates are available for the same target, we apply CI sequentially over the received estimates after gating to ensure they correspond to the same track (Sec.~\ref{sec:data_association}).
\subsection{Data Association}
\label{sec:data_association}
To associate incoming 3D detections with existing tracks, each UAV $j$ uses nearest-neighbor matching with Mahalanobis gating.
Let $\bm z_{j,k,t}\in\mathbb{R}^3$ be detection $k$ at time $t$ and $(\hat{\bm x}_{j,i,t|t-1},P^{x}_{j,i,t|t-1})$ the predicted state of track $i$.
Using the measurement model~\eqref{eq:track_meas}, we compute the innovation and its covariance
\begin{equation}
\begin{aligned}
\bm \nu_{j,k,i,t} &= \bm z_{j,k,t}-H\hat{\bm x}_{j,i,t|t-1},\\
S_{j,k,i,t} &= H P^{x}_{j,i,t|t-1} H^\top + R_{j,k,t}.
\end{aligned}
\label{eq:innov_short}
\end{equation}
and the squared Mahalanobis distance
$
d^2_{j,k,i,t}=\bm \nu_{j,k,i,t}^\top S_{j,k,i,t}^{-1}\bm \nu_{j,k,i,t}.
\label{eq:mahal}
$
Each detection is assigned to the track with minimum $d^2_{j,k,i,t}$ and accepted if $d^2\le\tau_{\mathrm{gate}}$.
Unmatched detections spawn new tracks; unmatched tracks are predicted forward and removed after a timeout or if their uncertainty exceeds a threshold.

\subsection{Multi-Target Assignment via Min-Cost Flow}
\label{sec:cmcf}
We assign UAVs to tracked containers using a CMCF formulation. At each allocation cycle,
the input is the fused target set $\{(\hat{\bm p}_i,P_i)\}_{i=1}^N$ and UAV positions $\{\bm r_j\}_{j=1}^M$, and the output is an
assignment set $\mathcal{A}_j$ for each UAV (up to $K$ targets), while enforcing a maximum range and inter-UAV separation.

\paragraph{Flow network.}
We build a directed graph with a source $sr$, UAV nodes $r_j$, target nodes $o_i$, and a sink $t$.
Edges $sr\!\to\! r_j$ have capacity $K$ (UAV capacity), edges $r_j\!\to\! o_i$ have capacity $1$ and cost $C_{ji}$ (assignment cost),
and edges $o_i\!\to\! t$ have capacity $1$ (each target assigned at most once). Solving the min-cost flow selects the UAV--target pairs.
\paragraph{Assignment cost.}
For a candidate pair $(j,i)$ we define the UAV--target distance
$
d_{ji}=\left\|\bm r_j-\hat{\bm p}_i\right\|.
\label{eq:dji}
$
We then compute an information-gain proxy using a simple range-dependent isotropic noise model:
\begin{align}
R(d) &= \sigma^2(d)\,I_3, \label{eq:R_of_d}\\
P_i^{+}(d_{ji}) &= \left(P_i^{-1}+R(d_{ji})^{-1}\right)^{-1}, \label{eq:Pi_plus}\\
\tilde{\Delta J}_{ji} &= \log\det(P_i)-\log\det\!\big(P_i^{+}(d_{ji})\big). \label{eq:info_gain_compact}
\end{align}
This favors assigning UAVs to targets whose uncertainty is expected to decrease most from the UAV's current position.
We then combine information gain, travel effort, and practical constraints:
\begin{align}
C_{ji}
&=
-\eta\,\tilde{\Delta J}_{ji}
+\beta\,d_{ji}
-\rho\,\mathbf{1}[i=i^{\mathrm{prev}}_j] \nonumber\\
&\quad
+\gamma\,\phi_j
+\kappa\,\mathbf{1}[d_{ji}>d_{\max}].
\label{eq:cost_simple}
\end{align}
where $\eta,\beta,\rho,\gamma,\kappa\ge 0$ are weights. The cost favors high expected uncertainty reduction ($\tilde{\Delta J}{ji}$), penalizes travel distance $d{ji}$, and includes a stickiness reward $\mathbf{1}[i=i^{\mathrm{prev}}j]$ to reduce assignment chattering. The term $\phi_j$ penalizes insufficient inter-UAV separation, and infeasible pairs with $d{ji}>d_{\max}$ are rejected. We solve the CMCF in real time via successive shortest paths, yielding the assignment sets $\{\mathcal{A}_j\}_{j=1}^M$.

\subsection{Hover Pose Selection}
\label{sec:hover_pose_selection}
Given an assigned UAV--target pair $(j,i)$, we select a hover pose $q^*_{j,i}$ from a discrete ring around the target by
trading off expected uncertainty reduction against travel distance.

\paragraph{Candidate viewpoints.}
We sample $L$ viewpoints on a ring of radius $r_h$ and altitude offset $h$:
\begin{equation}
q(\psi_\ell)=\hat{\bm p}_i+
\begin{bmatrix}
r_h \cos\psi_\ell\\
r_h \sin\psi_\ell\\
h
\end{bmatrix},
\qquad
\psi_\ell=\frac{2\pi\ell}{L},\ \ell=0,\dots,L-1.
\label{eq:hover_ring}
\end{equation}

\paragraph{Viewpoint scoring.}
At each candidate viewpoint we approximate the measurement noise as isotropic and range-dependent, compute the expected post-update covariance
$P_i^{+}(q)$ as in~\eqref{eq:info_gain_compact}, and score information gain with D-optimality:
\begin{equation}
\Delta J_i(q)=\log\det(P_i)-\log\det\!\big(P_i^{+}(q)\big).
\label{eq:dopt_gain}
\end{equation}
We then select the viewpoint that maximizes gain per travel distance from the current UAV position:
\begin{equation}
q^*_{j,i}
=
\arg\max_{\psi_\ell\ \text{feasible}}
\frac{\Delta J_i\big(q(\psi_\ell)\big)}
{\|\bm r_j-q(\psi_\ell)\|+\varepsilon},
\label{eq:hover_argmax}
\end{equation}
where feasibility enforces collision avoidance and inter-UAV separation. The selected pose $q^*_{j,i}$ is sent to the navigation controller for
\textsc{Tracking}.
\subsection{Surveillance and Tracking}
\label{sec:surveillance_and_tracking}

Each UAV operates in either \textsc{Surveillance} or \textsc{Tracking} mode. With no active assignment, it remains in \textsc{Surveillance} and follows the commanded patrol/waypoint plan using the navigation stack. When an assignment is produced (Sec.~\ref{sec:cmcf}), the UAV switches to \textsc{Tracking}, selects its most informative assigned target, and moves to the corresponding hover pose $q^*{j,i}$. Tracking ends once the assigned targets meet the termination criteria described below, after which the UAV returns to \textsc{Surveillance}.

\subsection{Tracking Termination Criteria}
\label{sec:tracking_termination_criteria}
After each allocation cycle, we prune ``completed'' targets from the candidate set.
A target track $i$ is marked as done if any of the following holds:
\begin{align}
\log\det P_i \;\le\; \tau_{\log\det},
\label{eq:term_logdet} \\
\max_{j\in\{1,\dots,M\}}
\ \max_{\psi_\ell \in \Psi^{\mathrm{feas}}_{j,i}}
\Delta J_i\big(q(\psi_\ell)\big)
\;\le\; \tau_{\Delta J},
\label{eq:term_dJ}
\end{align}
where $P_i$ is the current fused position covariance of target $i$,
$q(\psi_\ell)$ is a candidate hover viewpoint defined in~\eqref{eq:hover_ring},
$\Delta J_i(\cdot)$ is the expected D-optimality gain in~\eqref{eq:dopt_gain},
and $\Psi^{\mathrm{feas}}_{j,i}$ denotes the set of feasible ring angles for UAV $j$ around target $i$
(e.g., collision-free and respecting inter-UAV separation constraints).
$\tau_{\log\det}$ sets the maximum acceptable uncertainty: if $\log\det P_i \le \tau_{\log\det}$, the target is considered well localized.
$\tau_{\Delta J}$ sets a minimum information-gain threshold: if the best feasible viewpoint across all UAVs yields $\Delta J_i \le \tau_{\Delta J}$, the target is retired.

\section{SIMULATION RESULTS}
\label{sc sim-results}

The simulations were carried out on ROS 2 Jazzy, Ubuntu 24.04 (LTS) on Intel Core i7\textsuperscript{\textregistered}, with a system RAM of 32GB and NVIDIA RTX 1000 GPU. The code\footnote[1]{\url{https://github.com/MF-Ahmed/ship_drone}} and video demonstration\footnote[2]{\url{https://youtu.be/Vc9aGs_HevA}} is available. 

 \begin{figure}[H]
    \centering
      \subfloat[\label{fig:gz_a}]{%
           \includegraphics[height=3.2cm ,width=4.3cm]{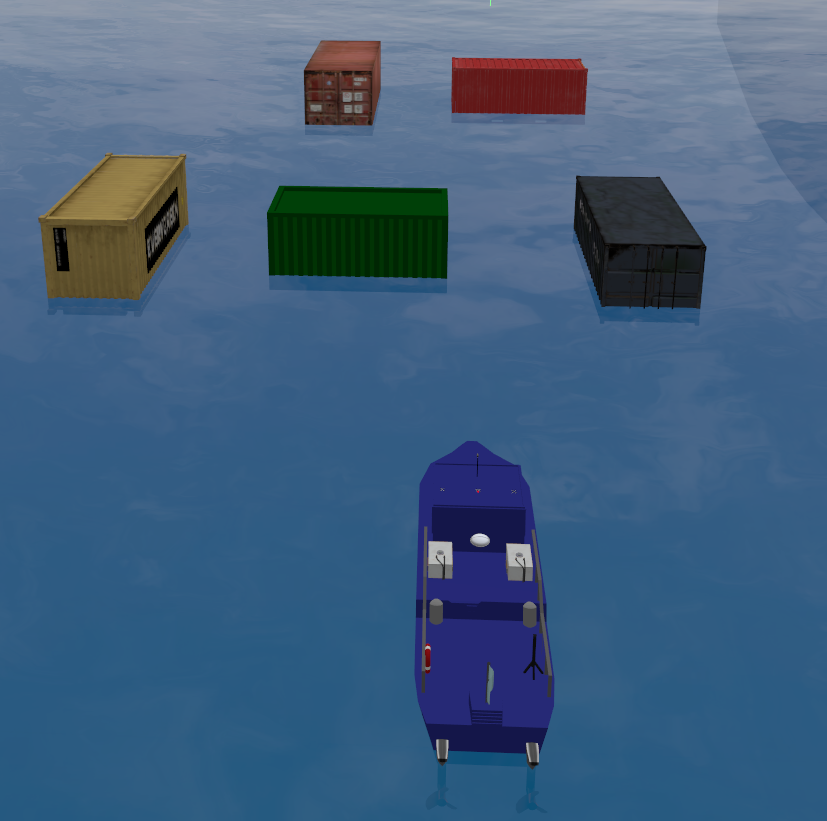}}
        \hfill
      \subfloat[\label{fig:gz_b}]{%
            \includegraphics[height=3.2cm ,width=4.2cm]{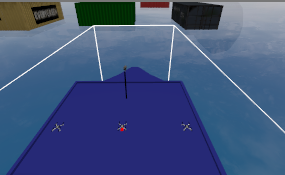}}              
    \caption{  Initial pos. of (a) Aquabot, containers and (b) drones.}

      \label{fig:gz}
    \end{figure}

We evaluate the proposed method in Gazebo simulation. We extend the Aquabot environment\footnote[3]{\url{https://github.com/oKermorgant/aquabot}}
 by adding three UAVs (\emph{drone~1}--\emph{drone~3}) and five floating transport containers (C1--C5). The UAVs are equipped with downward-facing RGB stereo cameras, GPS, and IMU sensors, while Aquabot carries an RGB camera, GPS, and an IMU. Fig.~\ref{fig:gz} shows the environment and the initial positions of the UAVs, containers, and Aquabot.

\begin{figure}[H]
    \centering
    \includegraphics[height=6.5cm ,width=8.5cm]{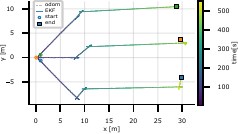}
    \caption{The mission trajectory, Drone1 (blue), Drone2 (orange) and Drone3 (green).}
    \label{fig:trajj}
\end{figure}

Fig.~\ref{fig:trajj} shows the UAV trajectories over the full 9-minute experiment, including initial and final positions. The mission comprises two phases (\emph{Path~1} and \emph{Path~2}). During \emph{Path~1} (0--220~s), the UAVs take off from Aquabot in \textsc{Surveillance}, spread out to maximize coverage, and switch to \textsc{Tracking} when containers are detected and assigned. They move to the selected hover-ring viewpoints around the targets and then return to \textsc{Surveillance}. During \emph{Path~2} (480--600~s), the UAVs follow a forward route and again transition to \textsc{Tracking} for newly assigned detections, visiting the corresponding hover-ring viewpoints.

\begin{figure}[H]
\centering
\setlength{\tabcolsep}{0.3mm}

\begin{tabular}{ccccc} 
\multicolumn{5}{c}{}\\
\cellimg{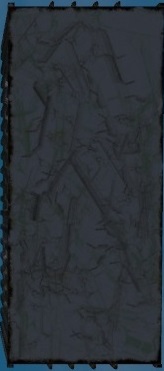}{\label{fig:rgb1}} &
\cellimg{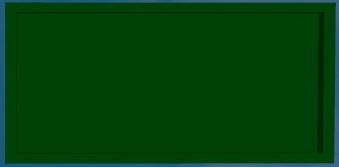}{\label{fig:rgb2}} &
\cellimg{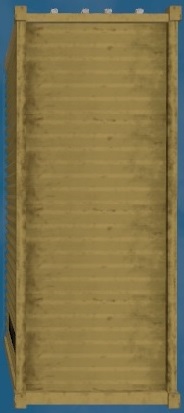}{\label{fig:rgb3}} &
\cellimg{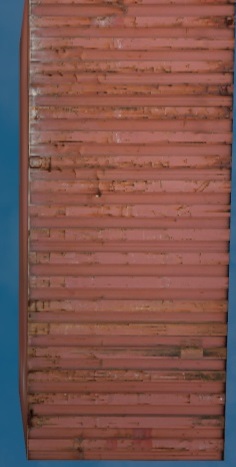}{\label{fig:rgb4}} &
\cellimg{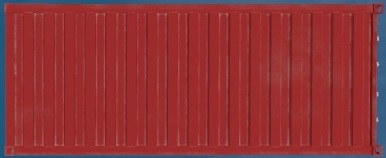}{\label{fig:rgb5}} \\

\multicolumn{5}{c}{} \\[-5mm]
\cellimg{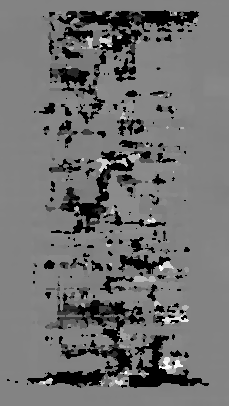}{\label{fig:disp1}} &
\cellimg{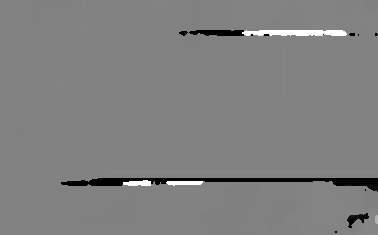}{\label{fig:disp2}} &
\cellimg{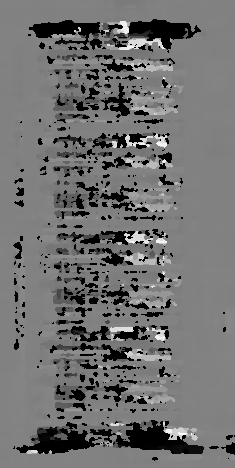}{\label{fig:disp3}} &
\cellimg{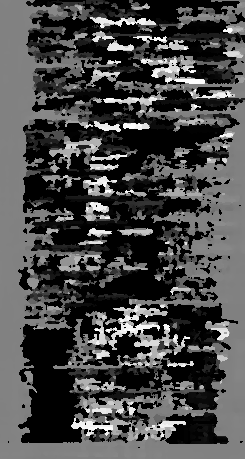}{\label{fig:disp4}} &
\cellimg{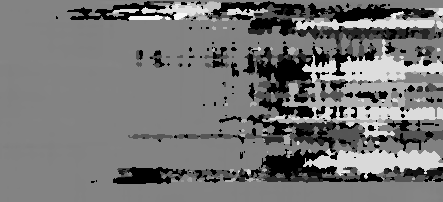}{\label{fig:disp5}} \\

\end{tabular}

\caption{Containers (C1--C5): RGB images (top row) and disparity images (bottom row).}
\label{fig:rgb_disp}
\end{figure}
Finally, Fig.~\ref{fig:rgb_disp} shows representative RGB images and the corresponding disparity maps. Disparity varies across containers due to surface-texture differences, leading to different depth contrast and structure within the container regions.

\subsection{Detection}
Table~\ref{tab:yolo_comparison} compares the baseline YOLOv8~\cite{yolo8} with our YOLOv8-RGBDis model. While precision/recall/F1 are near-saturated for both, our model achieves higher mAP@0.5 and mAP@0.5:0.95, indicating more reliable localization across IoU thresholds. Training is also more stable, and detections appear better calibrated in low-contrast scenes, supporting the benefit of incorporating disparity.

\begin{figure}[H]
    \centering
      \subfloat[\label{fig:map_all:1a}]{%
           \includegraphics[height=5.0cm ,width=4.2cm]{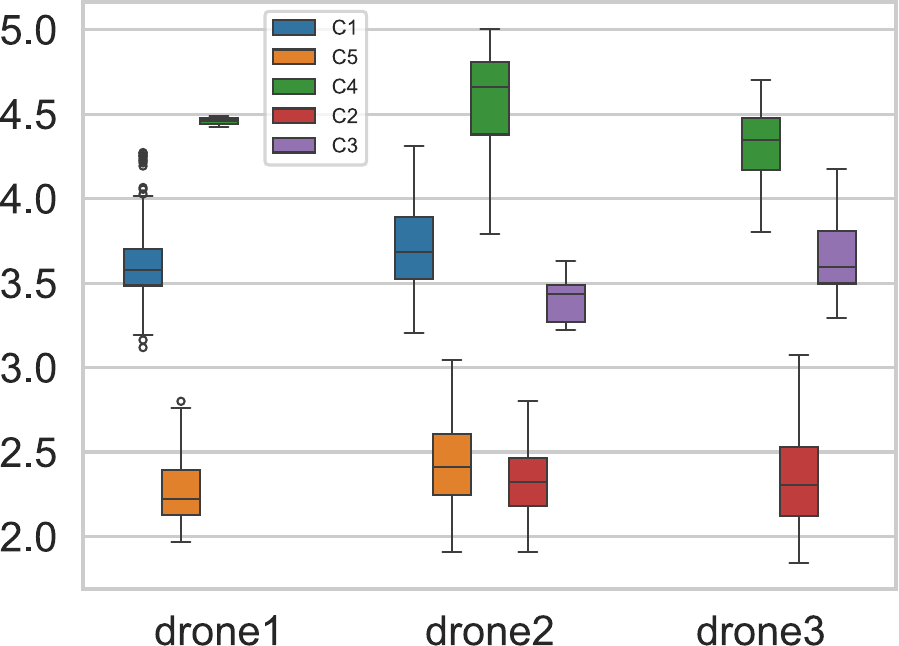}}
        \hfill
      \subfloat[\label{fig:map_all:1b}]{%
            \includegraphics[height=5.0cm ,width=4.2cm]{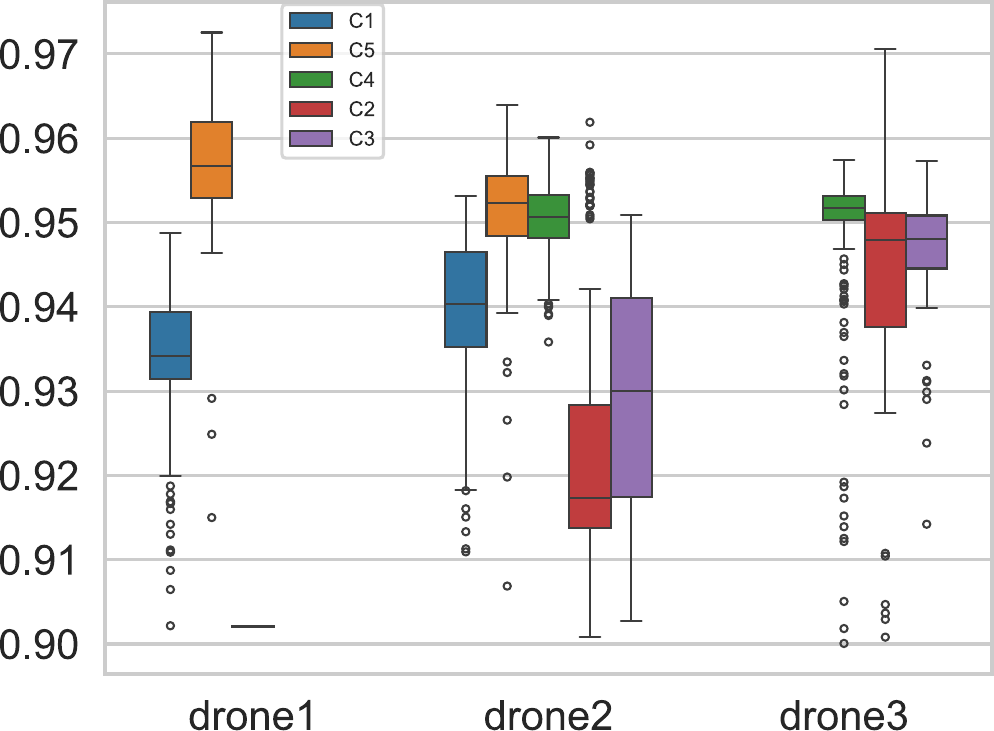}}              
\caption{(a) Detection error (m) and (b) confidence scores.}

      \label{fig: map_all}
    \end{figure}

\begin{table}[H]
\centering
\caption{Comparison between detection models.}
\setlength{\tabcolsep}{3pt} 
\renewcommand{\arraystretch}{1.1}
\resizebox{\linewidth}{!}{%
\begin{tabular}{lccccc}
\toprule
\textbf{Model} & \textbf{mAP@0.5} & \textbf{mAP@0.5:0.95} & \textbf{Prec.} & \textbf{Rec.} & \textbf{F1} \\
\midrule
\textbf{YOLOv8} & 0.985 & 0.952 & 0.998 & 1.000 & 1.000 \\
\textbf{Ours}       & \textbf{0.995} & \textbf{0.970} & 0.998 & 1.000 & 1.000 \\
\bottomrule
\end{tabular}%
}
\label{tab:yolo_comparison}
\end{table}
Fig.~\ref{fig:map_all:1a} reports localization errors over the full mission for all UAVs. The median error stays within $2.2$--$4.7\,\mathrm{m}$ depending on the container, with the lowest values for C2/C5 ($\approx 2.2$--$2.5\,\mathrm{m}$) and the highest for C4 ($\approx 4.3$--$4.7\,\mathrm{m}$). The dispersion is moderate (typical interquartile range $\approx 0.2$--$0.4\,\mathrm{m}$), indicating stable metric observations despite viewpoint changes.

Fig.~\ref{fig:map_all:1b} shows that detections are generally high-confidence and well calibrated. Median confidence values are typically $\approx 0.93$--$0.96$ across containers and UAVs, with small spread (IQR $\approx 0.005$--$0.015$). While a few outliers drop to $\approx 0.90$--$0.91$, the vast majority of detections remain above $\approx 0.93$, supporting their use as reliable EKF updates.

Overall, Fig.~\ref{fig: map_all} shows that the proposed perception front-end delivers accurate metric 3D measurements and well-calibrated confidence scores, supporting reliable EKF updates, robust tracking, and consistent multi-robot fusion throughout the mission.

\subsection{Tracking}
\label{sec:tracking}
We compare our method against two widely used multi-object tracking baselines: SORT~\cite{sort2016}, which associates 2D image-plane bounding boxes using IoU-based matching, and AB3DMOT~\cite{weng2020ab3dmot}, which associates metric 3D positions using distance-based matching and motion prediction. 
We report identity/continuity metrics (IDF1, IDSW, Frag) and 3D localization accuracy metrics (MedErr, RMSE, P95). 
IDF1 measures ID consistency over time (higher is better), IDSW counts identity switches (lower is better), and Frag counts track interruptions (lower is better). 
For 3D accuracy, we report the median position error, RMSE (penalizes large errors), and the 95th-percentile (P95) error to capture near worst-case performance (lower is better).

\begin{figure}[H]
\centering
  \subfloat[\label{fig:map_alll:1a}]{%
       \includegraphics[height=4.5cm ,width=4.31cm]{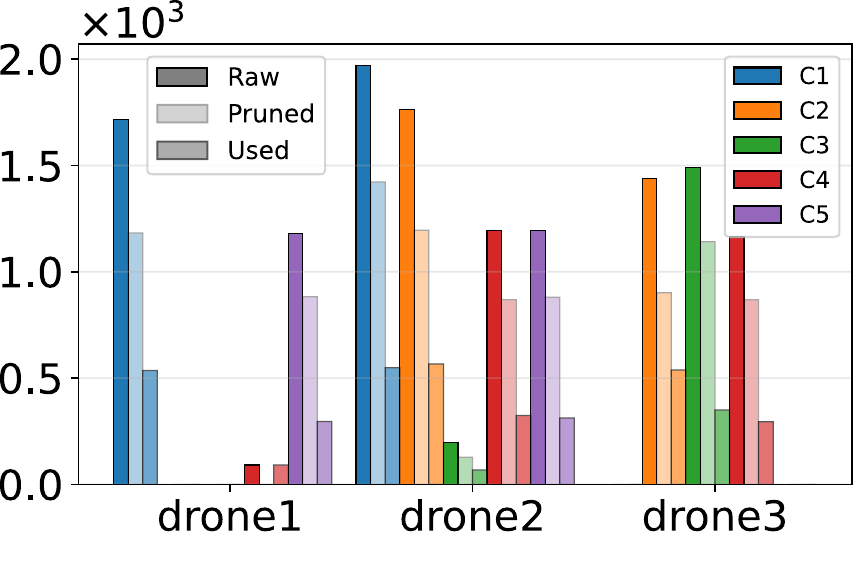}}
    \hfill
  \subfloat[ \label{fig:map_alll:1b}]{%
        \includegraphics[height=4.5cm ,width=4.31cm]{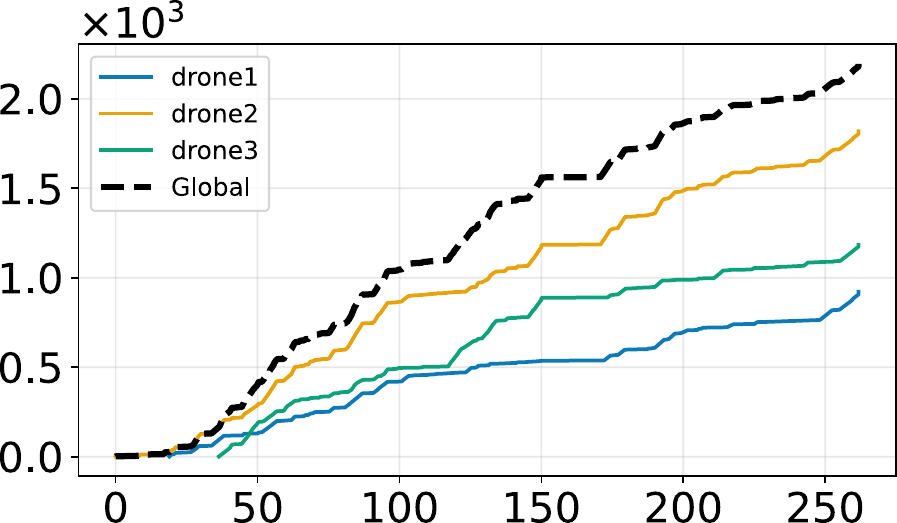}}              
    \caption{(a) Raw, pruned, and used tracks per drone; (b) cumulative number of pruned tracks over time (s).}
  \label{fig: map_alll}
\end{figure}

Table~\ref{tab:tracking_compact} summarizes tracking performance. We observe that 
our method achieves perfect identity consistency with no identity switches and low fragmentation, indicating stable target identities under intermittent observations. 
In 3D, it attains meter-level accuracy, comparable to AB3DMOT and substantially better than SORT. 
Although AB3DMOT achieves similar localization accuracy, it shows weaker identity continuity  and higher fragmentation, indicating more frequent track breaks. 
SORT performs markedly worse with large 3D errors, as expected from 2D-only association and higher sensitivity to clutter and imperfect depth lifting.
Figure~\ref{fig:map_alll:1a} illustrates the effect of Mahalanobis gating and track management: short-lived hypotheses are rejected, leaving a compact set of persistent tracks. 
This is particularly important in maritime imagery, where reflections and wakes can generate intermittent false positives that would otherwise proliferate and destabilize identities. 
Figure~\ref{fig:map_alll:1b} shows the cumulative impact of pruning over the mission: \emph{drone~1}, \emph{drone~2}, and \emph{drone~3} prune approximately $0.9\times 10^3$, $1.8\times 10^3$, and $1.2\times 10^3$ hypotheses, respectively, for a global total of $\approx 2.2\times 10^3$. 

Overall, combining metric stereo measurements with uncertainty-aware gating and EKF-based track management improves both identity continuity and 3D localization, yielding more reliable container tracks for fusion and task allocation.

\begin{table}[t]
\centering
\caption{Multi-object tracking performance.}
\label{tab:tracking_compact}
\resizebox{\columnwidth}{!}{%
\begin{tabular}{lcccccc}
\toprule
\textbf{Method} &  \textbf{IDF1}$\uparrow$ & \textbf{IDSW}$\downarrow$ & \textbf{Frag}$\downarrow$ & \textbf{MedErr}$\downarrow$ & \textbf{RMSE}$\downarrow$ & \textbf{P95}$\downarrow$ \\
\midrule
SORT & 0.390 & 46 & 142 & 12.709 & 12.621 & 14.328 \\
AB3DMOT & 0.899 & 0 & 13 & 3.431 & 3.259 & \textbf{3.875} \\
Ours  & \textbf{1.000} & \textbf{0} & \textbf{5} & \textbf{3.346} & \textbf{3.251} & 3.876 \\
\bottomrule
\end{tabular}%
}
\end{table}

We compare our CI-based fusion with fault-tolerant CI (FTCI)~\cite{FT}, a fault-tolerant variant of Covariance Intersection that remains robust to faulty or inconsistent inputs without requiring cross-correlation information. 
As shown in Table~\ref{tab:fusion_compact}, our method achieves lower MedErr/RMSE/P95, indicating slightly better accuracy and fewer large-error outliers. 
FTCI produces a marginally smaller $\overline{\log\det(P)}$ (i.e., a tighter covariance volume), but this does not translate into improved positional accuracy in our scenario. 
Overall, our CI fusion offers a better accuracy--uncertainty trade-off for the considered maritime setting while maintaining conservative covariance estimates under decentralized operation.


\begin{table}[t]
\centering
\caption{Fusion performance.}
\label{tab:fusion_compact}
\setlength{\tabcolsep}{3.5pt}
\renewcommand{\arraystretch}{1.05}
\begin{tabular}{lcccc}
\toprule
\textbf{Method} & \textbf{MedErr (m)$\downarrow$} & \textbf{RMSE (m)$\downarrow$} & \textbf{P95 (m)$\downarrow$} & $\overline{\log\det(P)}\,\downarrow$ \\
\midrule
Ours   & \textbf{3.523} & \textbf{4.126} & \textbf{6.830} & -5.715 \\
FTCI & 3.542 & 4.679 & 7.198 & \textbf{-5.818} \\
\bottomrule
\end{tabular}
\end{table}

\begin{figure}[H]
    \centering
    \subfloat[Pruning efficiency/time (s)\label{fig:goals:1a}]{%
        \includegraphics[height=3cm,width=4cm]{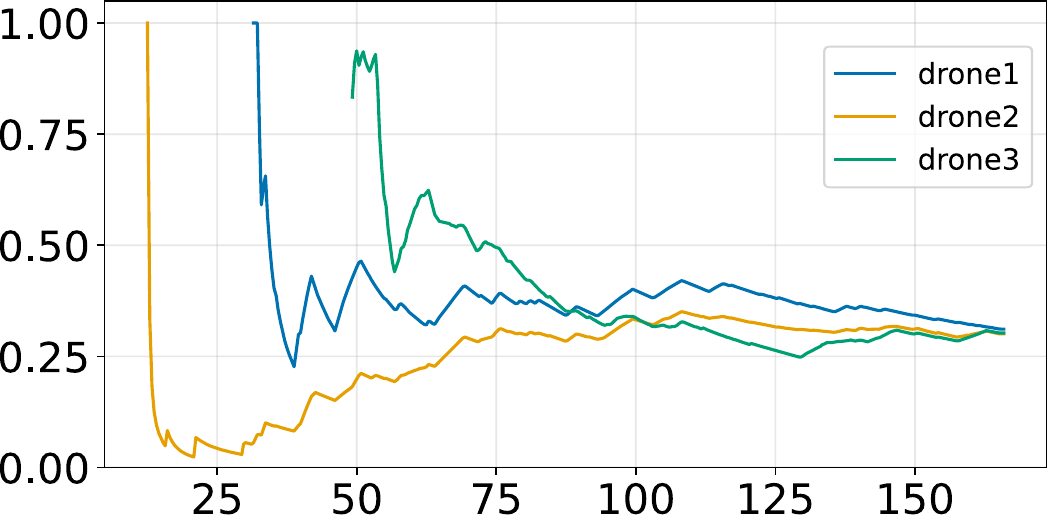}}
    \hspace{1mm}
    \subfloat[Raw, pruned tracks/time (s)\label{fig:goals:1b}]{%
        \includegraphics[height=3.2cm,width=4.3cm]{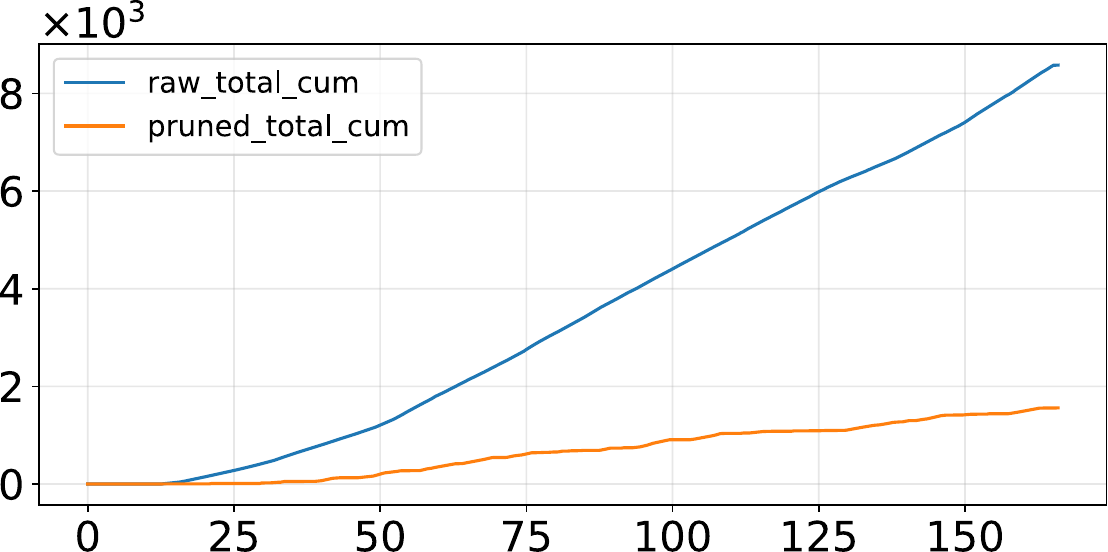}}
        \vspace{0.2mm} 
    \subfloat[Pruning efficiency/time (s)\label{fig:goals:1c}]{%
        \includegraphics[height=3cm,width=4cm]{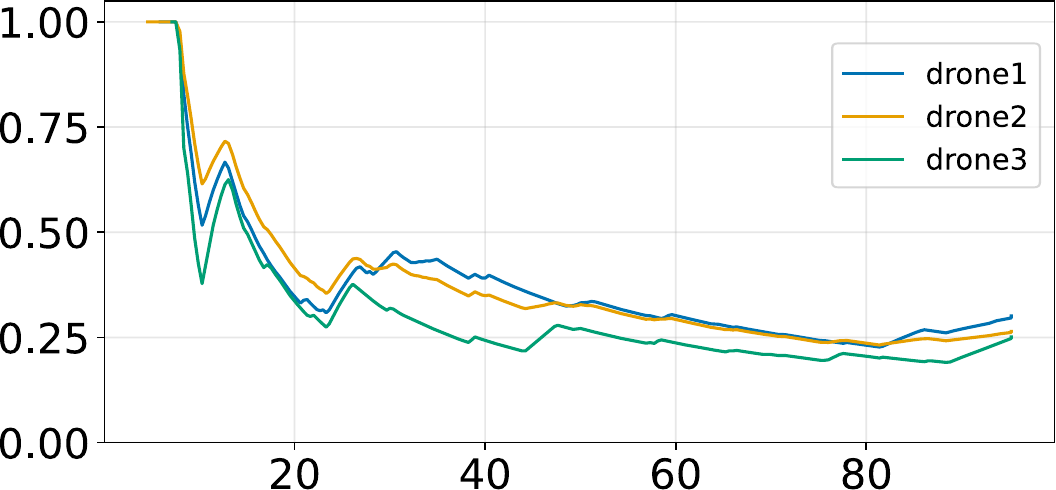}}
    \hspace{1mm}
    \subfloat[Raw, pruned tracks/time (s)\label{fig:goals:1d}]{%
        \includegraphics[height=3.1cm,width=4.3cm]{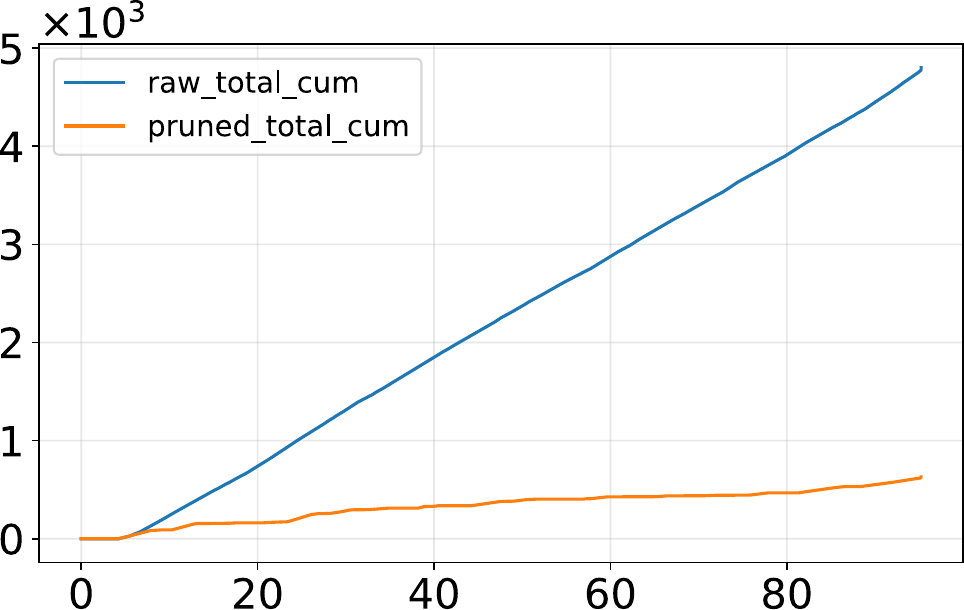}}

\caption{(a,b) Pruning efficiency and cumulative track evolution for \emph{Path~1}; (c,d) pruning efficiency and cumulative track evolution for \emph{Path~2}.}

    \label{fig:goals}
\end{figure}

For each UAV, we define pruning efficiency as the fraction of hypotheses rejected by gating:
\begin{equation}
\eta_{\mathrm{prune}}(t)=\frac{N_{\mathrm{pruned}}(t)}{N_{\mathrm{raw}}(t)}.
\end{equation}
In \emph{Path~1} (Fig.~\ref{fig:goals:1a}), $\eta_{\mathrm{prune}}$ stabilizes at $\approx 0.40/0.37/0.36$ for \emph{drone~1--3} (mean $\approx 0.38$). In \emph{Path~2} (Fig.~\ref{fig:goals:1c}), it drops to $\approx 0.26/0.23/0.17$ (mean $\approx 0.22$, $\sim$41\% lower), indicating fewer inconsistent hypotheses.

Cumulative counts follow the same trend: \emph{Path~1} reaches $\approx 8.6\times 10^{3}$ raw with $\approx 1.6\times 10^{3}$ pruned (Fig.~\ref{fig:goals:1b}); \emph{Path~2} reaches $\approx 4.8\times 10^{3}$ raw with $\approx 0.62\times 10^{3}$ pruned (Fig.~\ref{fig:goals:1d}). Overall, \emph{Path~1} generates more short-lived hypotheses, while gating keeps the retained set stable in both paths.

\subsection{Assignment}
\begin{figure}[H]
    \centering
    \includegraphics[height=8.0cm ,width=8.7cm]{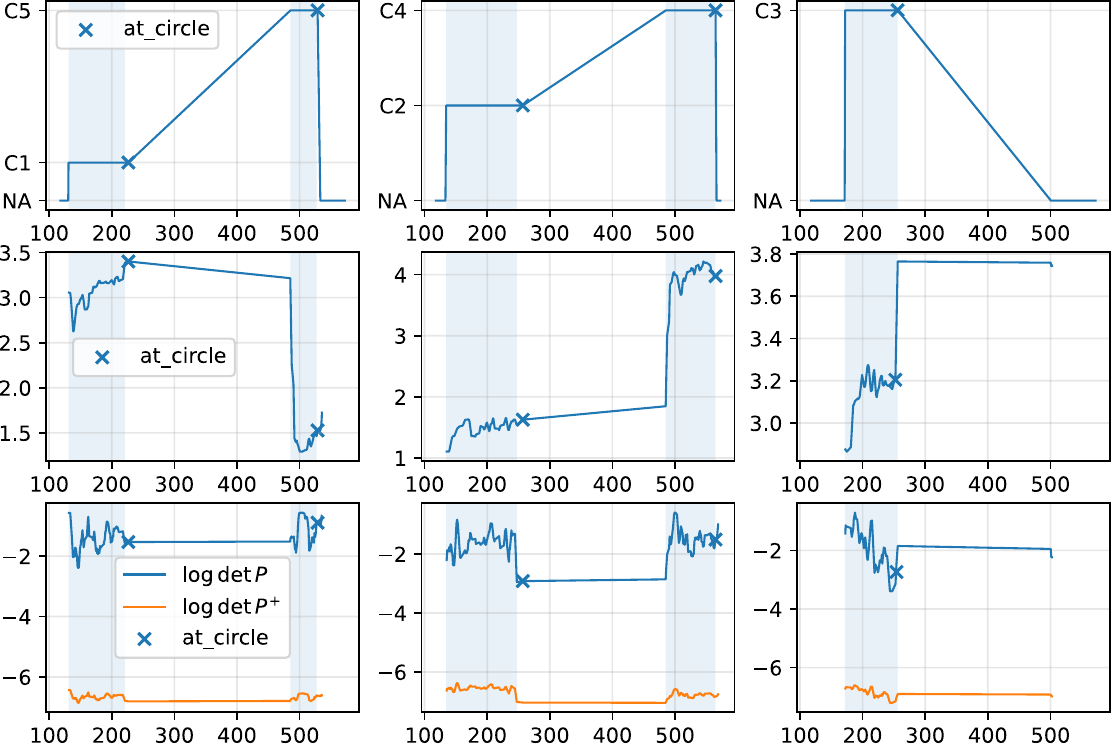}
    \caption{Assigned containers, position error, and uncertainty propagation of each drone over the full trajectory. }
    \label{fig:assigment}
\end{figure}

Fig.~\ref{fig:assigment} summarizes the CMCF assignment behavior (Sec.~\ref{sec:cmcf}). Rows~1--3 report, respectively, the assigned container, the XY position error, and the information-gain evolution, with columns corresponding to \emph{drone~1}--\emph{drone~3}. Pink and blue shading denote \textsc{Surveillance} and \textsc{Tracking}, and the mission is split into \emph{Path~1} and \emph{Path~2}. Row~1 shows that the allocator assigns C1--C3 in \emph{Path~1} and C5/C4 in \emph{Path~2} (\texttt{NA} indicates no assignment). In \emph{Path~2}, \emph{drone~3} remains unassigned because the available targets are already allocated to the other UAVs. Assignments are constant during \textsc{Tracking} and reset after handoff, reflecting clear mode switching. Row~2 shows bounded XY error: $\approx 1.3$--$3.4$\,m in \emph{Path~1}, and up to $\approx 3.9$\,m for \emph{drone~2} while tracking C4 in \emph{Path~2}. In contrast, \emph{drone~1} reduces error to $\approx 1.3$--$1.6$\,m when tracking C5, consistent with improved geometry after reaching the hover-ring viewpoint (Sec.~\ref{sec:hover_pose_selection}). Row~3 reports target uncertainty using the D-optimality score $\log\det(\cdot)$. Here, $P$ is the prior (predicted)
position covariance and $P^{+}$ is the posterior covariance after the measurement update. During \textsc{Tracking}, $\log\det P^{+}<\log\det P$, indicating consistent
uncertainty reduction after reaching the selected hover viewpoint.

Overall, Fig.~\ref{fig:assigment} shows that CMCF  keeps XY error bounded (often decreasing after reaching the hover ring), and steadily contracts target uncertainty until termination when further information gain becomes marginal.

\begin{figure}[H]
    \centering
    \includegraphics[height=4.0cm ,width=6.5cm]{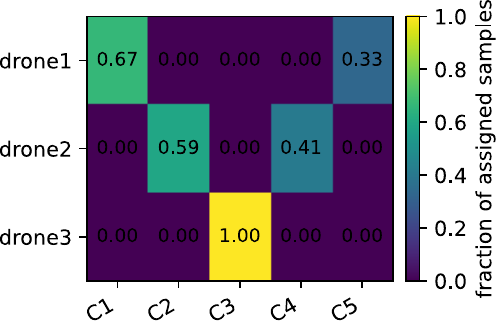}
    \caption{Drone and container assignment matrix. }
    \label{fig:assign_matrix}
\end{figure}

Fig.~\ref{fig:assign_matrix} reports the empirical assignment distribution (fraction of assigned samples) for the entire trajectory. \emph{Drone~1} is assigned primarily to C1 ($0.67$) and C5 ($0.33$), with negligible assignments to C2--C4. \emph{Drone~2} splits its assignments between C2 ($0.59$) and C4 ($0.41$), while \emph{drone~3} is exclusively assigned to C3 ($1.00$). Overall, the matrix exhibits a near one-to-one allocation pattern with limited overlap between UAVs, indicating that the allocator effectively partitions the targets and avoids redundant tracking while maintaining full coverage.

Overall, the simulation results support the proposed decentralized perception--tracking--assignment loop in a realistic maritime scenario. 
The perception front-end provides accurate metric observations despite reflective water effects, enabling reliable detections. 
Mahalanobis-gated EKF tracking with pruning suppresses short-lived false hypotheses and maintains a compact, stable set of tracks throughout the mission. 
Building on the fused tracks, the information-driven allocator with hover-ring viewpoint selection consistently directs UAV attention to the most uncertain containers, keeping position error bounded while steadily reducing track covariance.

\section{CONCLUSIONS}
\label{sc: conclusions}

We developed a decentralized multi-robot framework for maritime container detection and tracking, combining YOLOv8 and stereo 3D measurements, Mahalanobis-gated EKF tracking, and conservative Covariance Intersection fusion. In simulations, our approach achieved perfect identity continuity with low fragmentation and good accuracy. CI fusion further reduced large-error outliers while maintaining consistent uncertainty, and the information-driven allocator reliably assigned targets and selected informative hover viewpoints with modest communication. Future work will extend the system to real maritime trials under intermittent connectivity and sea-surface dynamics~\cite{future1}.

\section*{Acknowledgment}
This work was conducted within the framework of PerCoMa project funded by Carnot MERS institute, France.

\vspace{12pt}
\end{document}